\pdfoutput=1
\documentclass[11pt]{article}
\usepackage[a4paper,margin=2.4cm]{geometry}
\usepackage[T1]{fontenc}
\usepackage{lmodern}
\usepackage{microtype}
\usepackage{amsmath,amssymb}
\usepackage{mathrsfs}   %
\usepackage{graphicx}
\usepackage{booktabs}
\usepackage{array}
\usepackage[font=small,labelfont=bf]{caption}
\usepackage[numbers,sort&compress]{natbib}
\usepackage[hidelinks]{hyperref}
\usepackage{placeins}   %
\title{Collaborative AI agents for autonomous experimental research with solid-state quantum spins}
\author{El Mustapha Mansouri$^{*}$, Eikichi Kimura, Masato Koga and Keigo Arai$^{\dagger}$\\[0.4em]
\small School of Engineering, Institute of Science Tokyo, Yokohama, Kanagawa 226-8501, Japan}
\date{29 October 2026}

\begin{document}
\maketitle
{\renewcommand{\thefootnote}{\fnsymbol{footnote}}%
\footnotetext[1]{Corresponding author: \href{mailto:mansouri.e.2224@m.isct.ac.jp}{mansouri.e.2224@m.isct.ac.jp}}%
\footnotetext[2]{Corresponding author: \href{mailto:arai.k.835f@m.isct.ac.jp}{arai.k.835f@m.isct.ac.jp}}}

\begin{abstract}
An autonomous research laboratory could conduct an entire study, from formulating questions to preparing manuscripts. Artificial intelligence (AI) offers one route towards this goal. Recent AI systems take on several research roles, but may draw conclusions that the evidence does not support. Here, we present an autonomous research laboratory in which agents based on large language models assess one another’s work and can stop or redirect investigations. They can generate and reject research questions, require separate theoretical checks and conduct experiments designed to challenge their claims. We demonstrate these capabilities by investigating ensembles of quantum spins in diamond. The agents derived the minimum number of calibration settings needed to predict responses to a defined family of microwave pulse sequences. They checked the result computationally and designed experiments to distinguish competing predictions. An initial experiment agreed with the theoretical predictions, but repeated reference measurements showed that the signal was drifting during data collection. The agents therefore treated the result as inconclusive and revised the measurement protocol without relaxing the acceptance criteria. The revised experiment met those criteria and supported the theoretical predictions. This study demonstrates that collaborating AI agents can select research questions, derive theoretical results and test their predictions experimentally.
\end{abstract}

To conduct research autonomously, a system must decide what to investigate, how to test its ideas and what to conclude from the results. Robot scientists have generated hypotheses and tested them in the laboratory~\cite{king2004functional,king2009automation}, while autonomous laboratories have used experimental feedback to select and carry out further experiments~\cite{szymanski2023autonomous,burger2020mobile}. Systems based on large language models (LLMs) can also search the literature, propose ideas, plan experiments and analyse data~\cite{lu2026towards,gottweis2026accelerating,ghareeb2026multiagent}.

In recent studies, LLMs have challenged hypotheses, checked physical reasoning and revised interpretations in response to experimental evidence~\cite{isogawa2026agentic,yang2026endtoend,zeng2026socratic,gottweis2026accelerating}. A demanding test is whether agents can use this criticism to guide a study in which they develop a theoretical claim and design experiments to test it. Agents may complete every step of a study yet still draw a conclusion that the evidence does not support. In one evaluation, an AI Scientist reported performance improvements that were based on flawed experiments~\cite{beel2025evaluating}. We therefore ask whether agents can revise their experiments or conclusions when criticism reveals a flaw.

Scientific judgement becomes most consequential when it prevents an apparently promising investigation from proceeding. A plausible research question may already have an answer, and an elegant mathematical argument may rest on an unchecked assumption. For a proposed physical law, even a rigorous mathematical derivation cannot establish its empirical validity and relevance: its predictions must also withstand experiments designed to distinguish them from competing explanations. Organized scepticism requires these uncertainties to be confronted rather than merely acknowledged~\cite{merton1973normative}. Independent verification, precommitted predictions and acceptance criteria, and experimental controls provide safeguards against unjustified conclusions~\cite{nosek2018preregistration,simmons2011false,gelman2013garden}. The critical test is whether AI agents can enforce these safeguards, even when doing so means rejecting their own work.

Ensembles of nitrogen-vacancy (NV) centres in diamond provide a suitable setting for this test. Their optically readable spins can be manipulated through precisely controlled microwave-pulse sequences, while variations in spin properties create a nontrivial problem of predicting ensemble responses~\cite{doherty2013nitrogen,degen2017quantum,barry2020sensitivity}. This combination of experimental control and physical complexity allows mathematical predictions to be independently examined and tested against simpler calibration methods. The platform thus provides an opportunity to evaluate whether AI agents can formulate a meaningful question, establish a theoretical result and design an experiment capable of challenging that result rather than merely confirming it.

Here we present a collaborative AI laboratory that uses scientific criticism to guide an investigation from question selection to experimental assessment. Our objective is to determine whether collaborating agents can develop a substantive theoretical result and establish what experimental evidence supports its predictions. The agents pursued this objective by identifying the minimum calibration information needed to predict unmeasured responses of an NV spin ensemble, without reconstructing the unknown distribution of spin errors. The transition from theory to experiment required further collaboration: an experimentalist agent developed the control software and performed measurements, while a manager agent proposed redesigning the experiment rather than stopping after an agent serving as principal investigator rejected an initial design. During the research an experiment failed, the agents investigated the discrepancy and revised the acquisition procedure. The final measurements supported the calibration law developed by the proposer and PI. Together, the mathematical result and its experimental test demonstrate a scientific investigation in which AI agents collectively selected, challenged and substantiated a new physical claim.

\section{Collaborative AI laboratory for scientific discovery}
\begin{figure}[htbp]
\centering
\includegraphics[width=\textwidth]{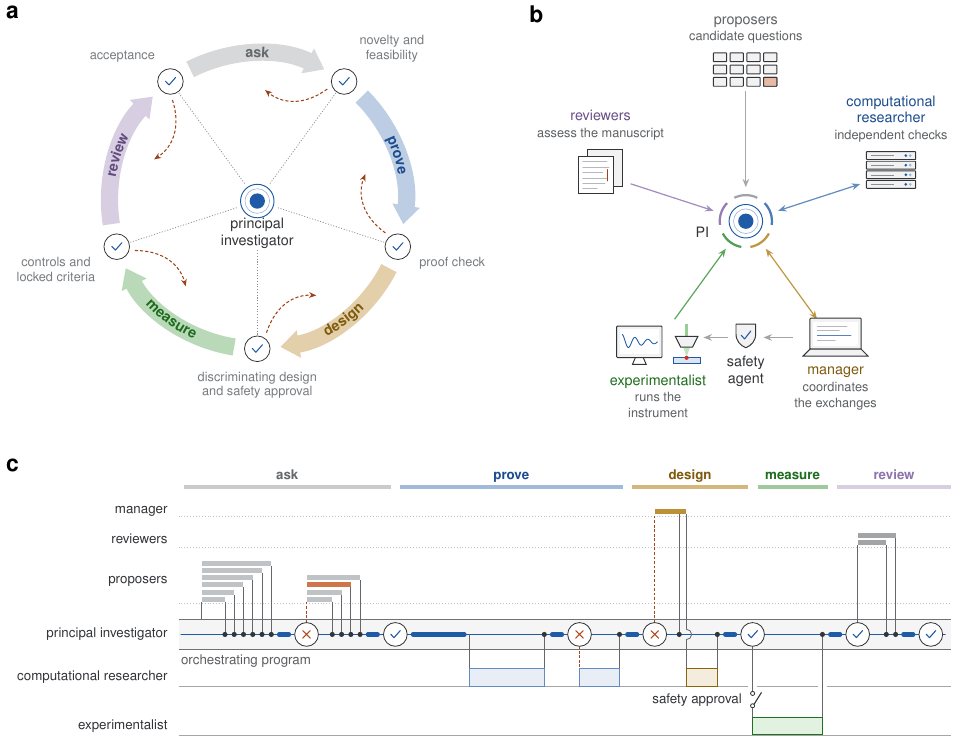}
\caption{\textbf{Scientific workflow and coordination of the collaborative AI laboratory.} \textbf{(a)} Five-stage research cycle with checkpoints overseen by the principal investigator (PI). Dotted lines indicate PI involvement; dashed arrows indicate rejection and revision. \textbf{(b)} Specialized AI agents and their exchanges of research outputs, with safety-agent approval required before experimentation. \textbf{(c)} Document flow during one pass through the API-orchestrated workflow, from left to right (not to scale). The orchestrator routes documents to the persistent PI conversation (blue). Dots mark document deliveries, circled ticks and crosses indicate PI decisions, and dashed links trace rejected work. Upper branches represent separate model calls; lower blocks represent coding-agent jobs.}
\label{fig:lab}
\end{figure}

To investigate whether scientific judgement can emerge through collaboration among AI agents, we developed a research system that distributes scientific responsibilities across specialized agents (Fig.~\ref{fig:lab}). The system connects research-question generation, theoretical analysis, computational verification and physical experimentation through exchanges of research outputs and critical feedback. Rather than automatically advancing results from one stage to the next, agents evaluate the work of their collaborators and can reject or request revisions when scientific criteria are not satisfied. By separating the generation and evaluation of scientific claims across agents with distinct contexts, this design aims to reduce self-confirmation bias and the reinforcement of erroneous assumptions across successive reasoning steps~\cite{huang2024large,li2026march}.

The laboratory comprised six principal agent roles: proposers, a principal investigator (PI), a computational researcher, an experimentalist, a manager and independent reviewers (Fig.~\ref{fig:lab}). The proposers independently generated candidate research questions, which the PI evaluated for scientific novelty, logical consistency, significance and experimental feasibility. After selecting a research question, the PI developed its theoretical foundations, while the computational researcher independently verified the mathematical results and examined the robustness of their underlying assumptions on a supercomputer. The experimentalist developed the instrument-control software and performed measurements on an NV-diamond platform. The manager coordinated exchanges between the agents and contributed to experimental planning, while independent reviewers assessed the resulting manuscript.

Scientific evaluation was integrated into the communication between agents rather than restricted to a final review. Research outputs were exchanged through written documents, and key decisions required explicit assessments before the investigation could proceed (Fig.~\ref{fig:lab}). Rejected proposals could be revised in response to criticism, mathematical claims were subjected to independent verification, and experimental designs could be reconsidered when they failed to distinguish the proposed theory from alternative explanations. Experimental predictions and acceptance criteria were fixed before the target measurements were analysed. These mechanisms allowed criticism to influence not only the interpretation of results, but also the direction and execution of the research itself.

Human researchers assembled the apparatus and configured the safety agent with preset operating limits. Within these boundaries, the AI agents carried out the calibration investigation through their respective roles, from research-question selection to experimental assessment.

\section{Critical evaluation and selection of research questions}
\begin{figure}[htbp]
\centering
\includegraphics[width=\textwidth]{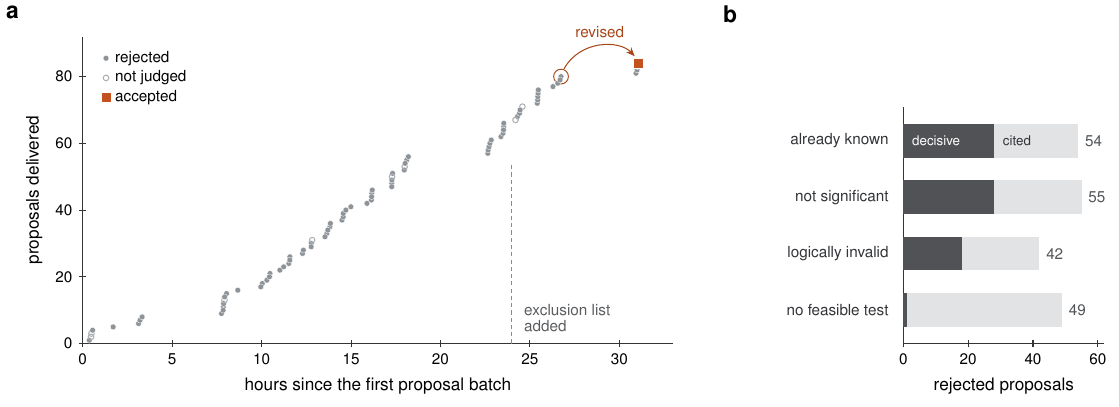}
\caption{How the accepted proposal emerged.
(a) Every proposal at its delivery time, marked by the PI's decision.
The dashed line marks when the repeatedly rejected idea families were listed in the proposers' instructions.
The ringed proposal was rejected but named the best seed, and its revision was accepted.
(b) Reasons for rejecting 75 proposals, in four categories: logically invalid (false or invalid
claims, overclaims), already known, not significant (no core theorem, trivial, renamed
mathematics, too narrow) and no feasible test.
Dark bars count the decisive reason and light bars every reason cited.}
\label{fig:search}
\end{figure}

We first examined how the laboratory selected a research question from independently generated proposals (Fig.~\ref{fig:search}). Each proposer received a description of the experimental platform and was tasked with identifying a mathematically substantive research question in machine learning or information science in which physics was essential, without access to the other proposals. The principal-investigator (PI) agent evaluated the proposals in batches, assessing their scientific novelty, logical validity, significance and experimental feasibility. Rather than selecting the most promising candidate by default, the PI could reject an entire batch and provide written criticism to guide subsequent proposals.

This selectivity proved consequential. Over approximately 30 hours, the PI rejected 75 proposals, accepting none during the first 17 batches (Fig.~\ref{fig:search}a,b). The rejections reflected recurring problems: previously established results, logical inconsistencies, insufficient scientific significance and the absence of feasible experimental tests. Moreover, independent proposers repeatedly converged on similar unsuccessful research directions. To overcome this limitation, the manager agent independently identified recurring rejected ideas and excluded them from subsequent proposal instructions. These ideas had accounted for 52 of the 56 earlier fresh proposals; following their exclusion, nine new proposals explored seven previously unrepresented research directions. The feedback therefore changed not only which proposals were rejected, but also the directions explored by the agents.

The eventual selection emerged through revision rather than acceptance of an initial proposal. The PI identified a previously rejected idea as a promising starting point and provided a critique that a new group of proposers used to reformulate the problem. The revised proposal replaced an approximation based on leading-order effects with a formulation retaining the full dependence on pulse sequences. This led the PI to select and refine a research question concerning the predictive information contained in measurements of an inhomogeneous spin ensemble: \textbf{what is the minimum number of calibration measurements required to predict the ensemble response to arbitrary phase-controlled pulse sequences?}

\section{Discovery and verification of an ensemble-calibration law}
The selected research problem concerned predicting the collective response of an inhomogeneous spin ensemble to phase-controlled microwave-pulse sequences. Individual spins experience different detunings and microwave-drive strengths, so the measured signal averages over an unknown distribution of pulse errors. The agents considered sequences of pulses with fixed duration, amplitude and shape but independently variable phases, assuming that each spin retained the same error parameters throughout a sequence. The selected proposal established a polynomial upper bound of $8L^2+6L+1$ independent response functions for sequences of up to $L$ pulses, but did not determine the exact number of calibrations required.

\begin{figure}[htbp]
\centering
\includegraphics[width=\textwidth]{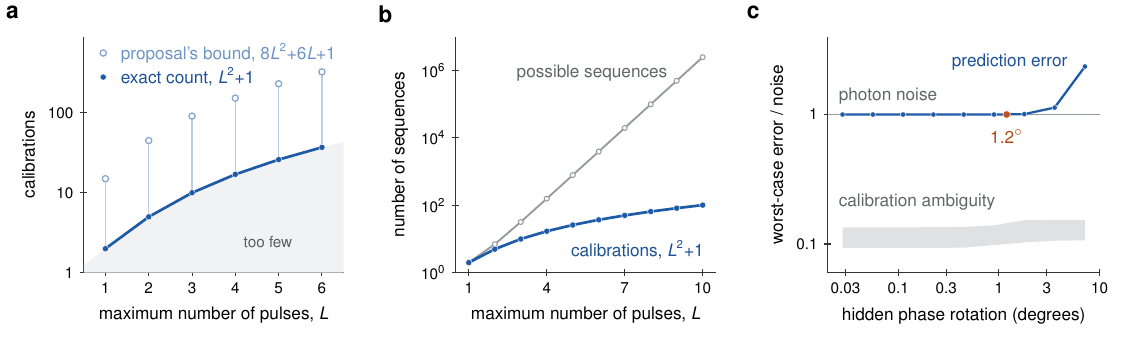}
\caption{Discovery and verification of a minimal ensemble-calibration law.
(a) The revised proposal supplied an upper bound of $8L^2+6L+1$ calibrations, which the principal-investigator agent refined to the exact necessary-and-sufficient count $L^2+1$ under persistent pulse errors.
(b) Calibration requirements compared with the number of possible sequences for five phase choices per pulse, with the first phase fixed.
(c) Certified prediction-error and calibration-ambiguity bounds under hidden phase rotations, relative to photon-counting noise.}
\label{fig:theory}
\end{figure}

During the subsequent theoretical investigation, the principal-investigator (PI) agent reformulated the problem and derived an exact result (Fig.~\ref{fig:theory}a). It proved that $L^2+1$ calibration measurements are both necessary and sufficient to predict the ensemble response to every phase-controlled sequence of at most $L$ pulses, independently of the distribution of persistent spin errors. The proof established that these measurements span the space of possible ensemble responses, even with an unknown common fluorescence background and contrast. Conversely, fewer calibrations cannot guarantee reconstruction, because distinct physically admissible ensembles can produce identical calibration measurements while differing on unmeasured sequences. The PI thus transformed the proposal's sufficient upper bound into an exact minimum, reducing the required count from the proposed bound of 91 to 10 for $L=3$. More generally, the calibration requirement grows quadratically with sequence length, whereas the number of possible pulse sequences grows exponentially for a fixed discrete set of phase choices (Fig.~\ref{fig:theory}b).

This result makes it possible to predict an entire family of control responses without reconstructing the unknown distribution of spin errors. With sufficient calibration, different microscopic ensembles can remain compatible with the measurements while giving identical predictions for every sequence in the specified family. The conceptual advance is the exact identification of the response information needed for this task and its equivalence to physically necessary calibration measurements. The resulting predictor takes the calibrated mean responses as inputs and returns mean responses for unmeasured phase-controlled sequences. The matching lower bound establishes a fundamental limitation: if calibration leaves a response direction unresolved, a more flexible predictor cannot guarantee its recovery for every admissible ensemble. The law therefore provides both a constructive way to select sufficient measurements and a criterion for when additional calibration is needed. The minimum $L^2+1$ refers to calibration settings with linearly independent responses; repeated acquisitions determine the precision of the measured means.

The mathematical result was then subjected to independent computational verification. A computational-researcher agent reproduced the symbolic checks and independently reconstructed the response-space structure, confirming dimensions of 26 and 37 for sequences of up to five and six pulses, respectively. The PI further challenged the assumption that pulse errors remain constant throughout a sequence by specifying a numerical stress test involving hidden phase rotations between pulses. Using the TSUBAME 4.0 supercomputer, the computational researcher generated 252,620 exact certificates bounding calibration ambiguity and prediction errors under these perturbations. Here, calibration ambiguity is the largest difference in target means between ensembles that agree on the short-sequence calibrations and remain statistically indistinguishable under the full calibration. For the specified six-pulse test, the certified prediction error remained at the photon-noise level for hidden phase rotations of approximately $1.2^\circ$ or less (Fig.~\ref{fig:theory}c). These calculations examined the robustness of the proposed law beyond its ideal assumptions, while its empirical validity remained to be established through measurements capable of distinguishing its predictions from simpler calibration methods.

\section{Collaborative design of a discriminating experiment}
\begin{figure}[htbp]
\centering
\includegraphics[width=\textwidth]{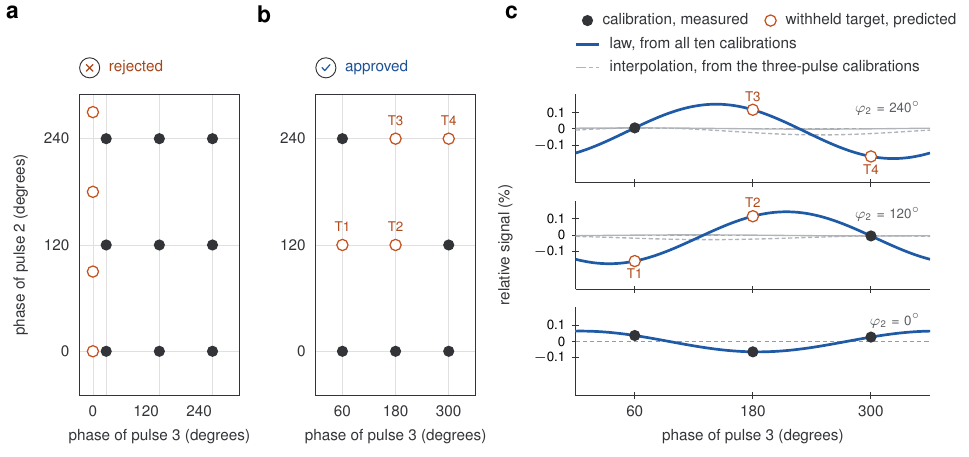}
\caption{Experimental design to distinguish the calibration law from interpolation.
(a) Rejected design using a complete grid of three-pulse calibrations, from which the withheld targets can already be determined by Fourier interpolation.
(b) Revised design combining five three-pulse calibrations with five shorter sequences to predict four withheld targets ($T_1$--$T_4$).
Filled symbols indicate calibration sequences and open symbols indicate withheld targets.
In both maps, rows give the phase of pulse 2 and columns the phase of pulse 3.
Both designs also include the same five shorter calibrations of zero to two pulses, which are not drawn.
(c) Measured signals of the final run along each row of (b), relative to the mean of the five three-pulse calibrations.
Dots are measured calibrations, whose standard errors are smaller than the dots.
The calibration law (blue) predicts each row from all ten calibrations, whereas the two interpolation-based alternatives (grey, solid and dashed) use only the five three-pulse calibrations.
Rings mark the withheld targets at the law's predictions, fixed before the target measurements were analysed; the measured targets are shown in Fig.~\ref{fig:result}c.}
\label{fig:discrim}
\end{figure}

The experimental objective was to test the predictive value of the calibration law using withheld sequence responses. The agents therefore sought a design in which the law and specified interpolation methods made different predictions, allowing the measurements to distinguish between them. The initial design combined a complete grid of nine three-pulse calibration sequences with five shorter sequences to predict four withheld targets (Fig.~\ref{fig:discrim}a). However, the nine-point grid already contained sufficient information to determine those targets through Fourier interpolation. Agreement between the predicted and measured signals would therefore not distinguish the proposed law from an ordinary interpolation method. The PI identified this limitation and rejected the design as an inadequate test of the theoretical result.

The rejection prompted further collaboration between the agents. The PI proposed a different approach that combined measurements from pulse sequences of different lengths, using five three-pulse calibrations together with five shorter sequences. The shorter sequences supply the information missing from the five three-pulse settings. The PI refined the phase settings and verified that these ten calibrations spanned the response space required by the law, while leaving four three-pulse targets that could not be determined by the competing interpolation methods (Fig.~\ref{fig:discrim}b). This cross-length design produced distinct predictions from the discovered law and simpler interpolation rules (Fig.~\ref{fig:discrim}c), providing an experimental test capable of distinguishing the competing explanations.

To ensure that the comparison could not be influenced by the observed targets, the PI established the prediction rules and acceptance criteria before data acquisition. The experimentalist was required to fix the numerical predictions obtained from the calibration measurements before analysing the withheld target data. The protocol also incorporated independent experimental controls and predefined conditions for classifying an outcome as supporting, contradicting or inconclusive, the last including any acquisition that failed a control. Before measurement, the experimental design was approved by a safety agent. Consequently, agreement with the theoretical prediction alone would not be sufficient to establish experimental support for the law.

\newpage

\section{Rejecting an apparently successful run and establishing \texorpdfstring{\mbox{experimental}}{experimental} support}
\begin{figure}[htbp]
\centering
\includegraphics[width=\textwidth]{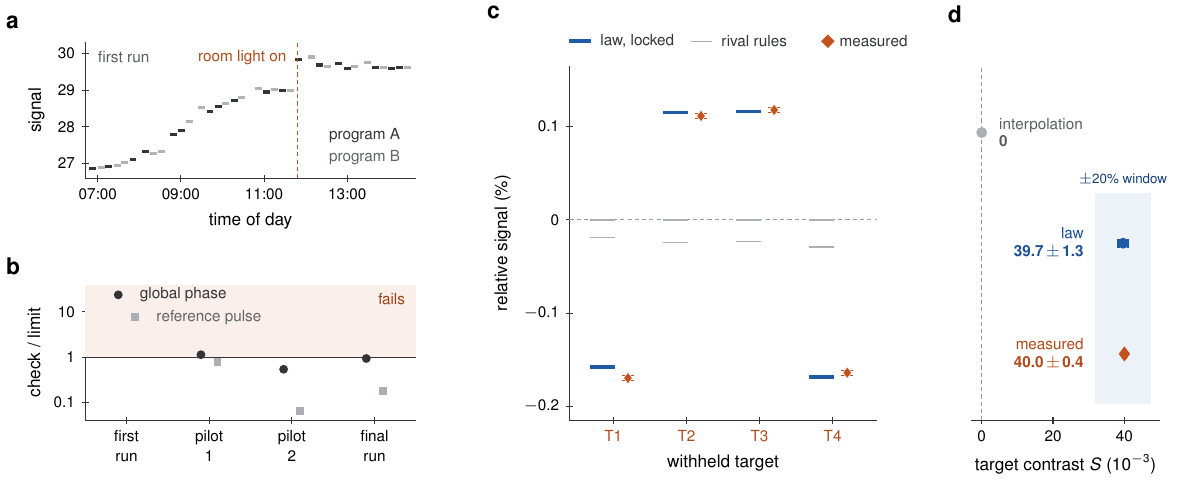}
\caption{Critical rejection and experimental support for the AI-derived calibration predictions.
(a) Signal of the first run in each 10-min sub-run, in weighted counts per shot, for the two acquisition programs; the dashed line marks 11:48, when the room light came on.
(b) The global-phase and reference-pulse checks of each acquisition, as a ratio to their preset limits; checks in the shaded region fail.
(c) The four withheld targets, relative to the mean of the five three-pulse calibrations: locked predictions of the calibration law (blue) and of the two interpolation-based rival rules (grey), and the final measurements (orange), whose standard errors are smaller than the symbols.
(d) The target contrast $S$ formed from the four targets in $10^{-3}$ weighted counts per shot; the shaded band marks the predefined tolerance of $\pm20\%$ around the law's locked prediction.}
\label{fig:result}
\end{figure}

The first main acquisition met all ten precommitted accuracy criteria, which assessed agreement for the four withheld targets and a contrast formed from those responses (Methods). However, the global-phase and reference-pulse controls failed (Fig.~\ref{fig:result}b). These controls tested whether applying the same phase shift to every pulse left the measured response unchanged. The principal-investigator (PI) agent therefore classified the run as inconclusive.

The experimentalist traced the failed controls to changes in the signal during acquisition. The signal rose by 7.6\% over the morning and showed a further 2.9\% step at 11:48 a.m., coinciding with a laboratory member switching on the room lights (Fig.~\ref{fig:result}a). The experiment alternated between two programs containing different global-phase replicas. Because these programs sampled the drifting signal at different times, measurements expected to be equivalent acquired systematic offsets.

The agents revised the acquisition protocol without relaxing the acceptance criteria. The experimentalist shortened the alternation between programs and tested different sequence orders in pilot acquisitions (short runs used to test adjustments to the acquisition procedure). The first pilot still failed the phase controls; a second, with phase switching within a single program, passed. The PI retained the inconclusive verdict for the original run and specified a new six-hour acquisition with randomized sequence ordering and additional diagnostics of the null pair. The manager conveyed the revised protocol to the experimentalist, who implemented it.

The final acquisition yielded 21,600 valid one-second records and passed all ten accuracy criteria and experimental controls (Fig.~\ref{fig:result}b). Predictions from the calibration measurements were fixed before the withheld targets were analysed. The four withheld responses were summarized by the target contrast $S=(-T_1+T_2+T_3-T_4)/4$, which cancels the common fluorescence background. The measured target contrast was $(40.0\pm0.4)\times10^{-3}$ weighted counts per shot, consistent with the law's prediction of $(39.7\pm1.3)\times10^{-3}$, whereas the interpolation alternatives predicted zero (Fig.~\ref{fig:result}c,d). These measurements support the transfer of information from the shorter calibration sequences to the four withheld three-pulse responses within the predefined accuracy tolerances.

\FloatBarrier
\section{Discussion}
Recent debate about AI scientists concerns whether automated research can produce conceptual insight and reliable scientific judgement~\cite{lu2026towards,nature2026why}. Our results demonstrate both in a physical investigation. Critical evaluation shaped the research question, strengthened the mathematical theory and determined which experimental evidence could support a conclusion. The principal-investigator agent rejected an experimental design because conventional interpolation could predict the same targets. It subsequently retained an inconclusive verdict for an acquisition that met all ten precommitted accuracy criteria but failed independent experimental controls. The agents revised the acquisition protocol without relaxing the acceptance criteria. These decisions demonstrate that collaborating AI agents can distinguish predictive agreement from sufficient scientific support and enforce that distinction when evaluating their own work.

The calibration law provides a concrete example of conceptual progress during the investigation. The PI reformulated calibration in terms of the dimension of the ensemble-response space and proved that $L^{2}+1$ measurements are both necessary and sufficient for sequences of at most $L$ pulses under persistent pulse errors. Establishing necessity required identifying physically admissible ensembles that agree on an insufficient set of calibrations yet differ on unmeasured sequences. This reasoning transformed a sufficient construction into an exact characterization of the information required for prediction. The experimental measurements supported this predictive capability on the four withheld targets within the predefined tolerances, distinguishing the law’s predictions from those of the two specified interpolation methods. Together, the theoretical result and its experimental test show how an AI laboratory can identify a precise information requirement, turn it into a measurable prediction and assess the evidence supporting that prediction.

Solid-state spin defects provide a particularly promising platform for extending this approach. Nitrogen-vacancy centres combine optical spin initialization and readout with programmable microwave control, allowing many experimental protocols to be executed without manual intervention once the apparatus is operational~\cite{isogawa2026agentic,doherty2013nitrogen,degen2017quantum,barry2020sensitivity}. Their sensitivity to control imperfections and environmental fluctuations also provides demanding tests of quantum-control theories. Such systems could enable AI agents to investigate new pulse sequences, calibration strategies and sensing protocols through repeated cycles of theoretical prediction and physical measurement. Similar approaches may be applicable to other quantum platforms.

Extending the laboratory towards greater autonomy will require improvements in both the agent architecture and the experimental infrastructure. Larger groups of parallel proposers could broaden the search for research questions, while more specialized agents could extend safety assessment or coordinate work across laboratories. Limited diversity in generated research ideas has been identified in evaluations of LLM research agents~\cite{si2025can}. Here, the manager used the record of rejected proposals to revise the search strategy, after which the proposers explored previously unrepresented research directions. This adaptation demonstrates that an agent-led investigation can use accumulated criticism to change the scientific questions it pursues. Connecting multiple experimental setups, including cryogenic platforms, would further expand the range of testable physical phenomena. Such extensions would also require more reliable instrument diagnostics, coordination and safety mechanisms as agents gain additional experimental degrees of freedom.

A broader challenge is therefore to develop experimental testbeds in other branches of science that allow AI-generated hypotheses to be tested with comparable independence and rigour without human involvement. Greater scientific autonomy will depend not only on improvements in AI reasoning, but also on instruments that make theoretical predictions experimentally accessible and competing explanations distinguishable. As these capabilities develop, the role of human researchers may shift towards identifying consequential questions, expanding experimental possibilities and critically assessing the significance of automated discoveries. The long-term opportunity is not simply to automate scientific work, but to extend the range of questions that can be investigated and answered with reliable evidence.

\section{Methods}

\subsection{Agent implementation and human contributions}
We implemented the laboratory using model conversations and coding-agent sessions configured for different research roles. The principal investigator (PI) used GPT-6 at the Pro reasoning setting, with Extra High for reviews. Proposal generation used GPT-5.6 extra high. The computational researcher ran GPT-6 Astra at maximum reasoning effort through Codex, using CPUs and GPUs on the TSUBAME 4.0 supercomputer. The experimentalist used Claude Fable 5 through Claude Code on the instrument computer, and the manager used Claude Opus 5 on a separate computer with browser and SSH access to the experimentalist and the computational researcher on the supercomputer.

Mechanisms explored in earlier projects were incorporated into the exclusion instructions for subsequent proposal searches to prevent new proposals from overlapping with previously explored research directions. Each role received instructions specifying its scientific task, expected output and evaluation criteria.

For the NV investigation, human researchers assembled the apparatus and supplied a description of its capabilities. During setup, they configured and checked the safety agent using preset operating limits, including maximum voltage and power.

\subsection{Context allocation, document exchange and review}
Agents exchanged proposals, manuscripts and research reports as written documents relayed by the manager. Information was supplied according to each agent’s task. Fresh proposers received a short description of the experimental platform and were not shown one another’s proposals. Each proposal batch was assessed in a new PI conversation. After selection, the PI continued that conversation through theoretical development, manuscript preparation and computational test design. A separate feasibility reviewer received the manuscript and full apparatus specification without the PI’s preceding exchanges.

An internal manuscript carried the research question, assumptions and mathematical derivations between agents. The computational researcher received its source files together with the PI’s test specification, and returned verification results and a revised manuscript. Its working directory persisted across tasks. The experimentalist received the experimental protocol and task instructions, maintained local notes and logs, and returned laboratory reports. Raw photon counts remained on the instrument computer; the PI and computational researcher assessed the measurements from these reports.

Assessments at decision points ended with an explicit verdict and written reasons. Requests for revision specified numbered changes for the agent responsible for the task. The decisive computational experiment underwent one PI review followed by one correction pass by the computational researcher. Final manuscript reviewers started in new conversations and received only the manuscript, without the preceding agent exchanges, source code or raw data. They returned a referee report, recommendation and confidence score. After receiving the reviews, the PI judged the experimental evidence sufficient and instructed the computational researcher to add more information in the manuscript as requested by the reviewers.

\subsection{Proposal generation and adaptive selection}
Each proposer was instructed to formulate one theoretical contribution to information science in which physics was essential and that could be applicable to NV centres. The selected proposal met this brief as a learning problem: it asked how many calibration measurements are necessary and sufficient to predict the responses of an exponentially large family of control sequences, a question of identifiability and sample complexity whose answer is set by the physics of persistent pulse errors. Proposals had to address a significant problem, identify a feasible theoretical approach and include at least one central prediction specific to the idea and testable on the available NV platform. Numerical validation and empirical benchmarks were excluded at this stage. Proposers were instructed to develop candidates before checking the literature, assess novelty by the underlying mechanism, and exclude directions adopted from published future-work or open-problem discussions.

Fresh batches launched ten proposer conversations to avoid overloading the context of the PI. Each fresh batch was forwarded to the PI once five proposals had arrived, and the remaining proposers were stopped. A separate deduplication stage grouped proposals by central mechanism and retained one representative from each group. The PI required proposers to return the strongest idea reached, with remaining uncertainties stated. The PI assessed novelty, logical validity, significance, theoretical feasibility and experimental testability, and could reject an entire batch.

Proposal instructions excluded mechanisms explored in earlier projects. After twelve consecutive assessments without a winner, the manager added exclusions derived from repeatedly rejected mechanisms in the PI’s verdicts. Revision batches launched six proposers, each receiving a rejected proposal verbatim and the PI’s full assessment of its batch. They were instructed to address decisive objections and reassess novelty, with permission to change the mechanism if the original could not be sustained. Returned revisions were not used as seeds for further revision batches.

The assessment criteria and literature checks were retained, and the selection instruction was to choose and strengthen the best proposal while stating remaining uncertainties. Selecting none was permitted only if no candidate was novel, feasible or impactful. The PI was instructed to distinguish justified claims from plausible statements and outstanding proof obligations. Before manuscript preparation, it compared the contribution with the five closest works, specifying what each established and what remained new.

\subsection{Spin-response model and calibration construction}
The PI and the proposer who originated the research considered pulses with fixed duration, amplitude and shape and independently programmable phases. In the two-level model, a pulse of phase $\phi$ was represented by
\[
U_\phi(a)=
\begin{pmatrix}
a & \sqrt{q}\,e^{-i\phi}\\
-\sqrt{q}\,e^{i\phi} & \bar a
\end{pmatrix},
\qquad q=1-|a|^2,
\]
where $a$ encodes the pulse action on an individual spin and $\bar a$ denotes complex conjugation. Each spin retained the same $a$ throughout a sequence. The ensemble was described by an arbitrary unknown probability distribution $\nu$ on an admissible domain $K\subseteq\{a:|a|\leq1\}$ with nonempty two-dimensional interior.

For a sequence $w=(\phi_1,\ldots,\phi_\ell)$ in time order, with $\ell\leq L$, the return probability and mean fluorescence signal were
\[
p_w(a)=\left|\langle0|U_{\phi_\ell}(a)\cdots U_{\phi_1}(a)|0\rangle\right|^2,
\qquad
\mu(w)=B+\kappa\int_K p_w(a)\,\mathrm d\nu(a).
\]
The empty sequence had $p_{\varnothing}=1$. The distribution $\nu$, unknown offset $B$ and nonzero contrast $\kappa$ were common to all sequence settings.

The PI and the proposer defined $\mathcal F_L$ as the real span of the response functions for sequences of at most $L$ pulses. Uniform identification of every target in this class required calibration responses spanning $\mathcal F_L$, whose dimension was $d_L=L^2+1$. The general construction used the five phases $2\pi k/5$, $k=0,\ldots,4$, to resolve the phase Fourier components and selected $d_L$ linearly independent sequence responses as a basis. For each target, expansion in this basis determined real coefficients $h_j(w)$ satisfying
\[
p_w(a)=\sum_{j=1}^{d_L}h_j(w)p_{b_j}(a),
\qquad
\sum_{j=1}^{d_L}h_j(w)=1.
\]
Target predictions were therefore computed as
\[
\widehat{\mu}(w)=\sum_{j=1}^{d_L}h_j(w)\widehat{\mu}(b_j),
\]
where $\widehat{\mu}(b_j)$ denotes the measured mean for calibration sequence $b_j$. The count $d_L$ refers to distinct calibration settings; repeated acquisitions determine their precision. The specific laboratory calibration design is described below.

For the theoretical photon-noise analysis, acquisitions were modelled as independent Poisson counts with means bounded by $M$. A determinant-maximizing acquisition design with $N\geq d_L$ observations, combined with least-squares estimation, gave worst-case mean-squared prediction error at most $Md_L/(N-d_L+1)$.

\subsection{Independent computational verification and robustness analysis}
The computational researcher reproduced the symbolic checks and independently reconstructed response-space coordinates for $L=5$ and $L=6$. A complementary stress test, specified by the PI, examined whether departures from persistent pulse errors could remain statistically hidden in calibration while changing target predictions. This test fixed
\[
L=6,\qquad
K=[3/5,4/5]+i[-1/5,1/5],\qquad
B=1,\qquad \kappa=1/100.
\]
Five phase settings with rational sine and cosine values were used, with the first pulse phase fixed. The design comprised 192 calibration sequences, including every sequence through three pulses, and 3,715 targets. A hypothetical budget of $8.8\times10^{10}$ Poisson acquisitions, with approximately one photon per acquisition, set the statistical scale.

Perturbations consisted of at most one hidden rotation about each spin’s $z$ axis, with angles $+\delta$ and $-\delta$ occurring with equal probability. Rotations after pulse three, four or five were considered separately, together with an unperturbed case. Calibration ambiguity was defined as the largest target-mean separation between ensembles that agreed on all short-sequence means and were statistically indistinguishable under the full calibration. The indistinguishability criterion required an optimal discrimination success probability of at most 55\%.

Candidate ensemble pairs were sought numerically, and bounds were certified using exact rational arithmetic. The domain $K$ was subdivided into rectangles, on each of which polynomial inequalities were checked using the minimum and maximum Bernstein coefficients. This procedure certified the inequalities across the continuous parameter domain for each specified test case. A Berry--Esseen bound on sums of photon counts supplied a necessary condition for the indistinguishability criterion and was used to tighten the ambiguity bounds.

Separate certificates bounded the calibrated predictor’s squared bias and variance. Their sum defined the mean-squared prediction error, whose square root was expressed relative to $\sigma_M$, the predictor’s photon-noise standard deviation. Certification covered every target, the three rotation positions and the unperturbed case at twelve evaluated rotation strengths. Seven exact interval halvings refined the crossing of the photon-noise reference. A control retained only the 32 calibration means through three pulses and sought ensembles agreeing on those means but differing on a six-pulse target.

\subsection{Experimental design and prediction locking}
\label{sec:mdesign}
Experimental proposals underwent separate assessments of apparatus feasibility and scientific discrimination. The feasibility reviewer evaluated whether the available apparatus could test a central prediction, and the PI assessed whether the proposed measurements distinguished that prediction from specified alternative models.

The final design combined five three-pulse calibration sequences with five sequences of at most two pulses, including the empty sequence. Four additional three-pulse settings served as withheld targets. Phases were selected in multiples of $60^\circ$, and an exact rank calculation verified that the ten calibration responses spanned the $L=3$ response space. The PI derived a $4\times10$ transfer matrix $H$, with each row summing to one, to obtain target predictions as $\widehat{\mathbf T}=H\mathbf C$. Denoting calibration and target means in weighted counts per shot by $C_0,\ldots,C_9$ and $T_1,\ldots,T_4$, respectively, the test compared the measured contrast
\[
S=\frac{-T_1+T_2+T_3-T_4}{4}
\]
with the calibration prediction
\[
P=\frac34(C_0-C_1+C_6)-\frac14(C_2+C_3+C_4).
\]
Both combinations cancel the common fluorescence background.

Two interpolation baselines were specified. A nine-term Fourier interpolant fitted the five three-pulse calibration means by minimizing the squared coefficient norm. A six-term alternative incorporated the symmetry of the repeated pulse. Both predicted $S=0$ for any calibration record. The protocol additionally required a nine-parameter model of coherent ensemble evolution to be fitted to all ten calibration means.

A digital twin created by the computational researcher evaluated the decision rule using 20,000 simulated runs per case, including intended operation, a target phase misprogrammed by $60^\circ$, and cases at the accuracy-tolerance boundaries. The laboratory packet contained the protocol, data format and analysis code, and underwent a separate review before execution. Its feasibility assessment used an approximately 12-hour budget on the HPC, including calibration, without relaxing the prescribed thresholds.

Before acquisition, the PI fixed $H$, the definition of $P$, the accuracy tolerances, controls and outcome criteria. Initial analysis was restricted to the ten calibration means and their covariance. The protocol required the numerical law predictions, $P$, both interpolation predictions and the physical fit to be locked and hashed before target analysis. For the final acquisition, the documented prediction lock occurred after data collection and before any target mean was computed. The statistical outcome rules are specified below.

\subsection{NV apparatus and fluorescence readout}
Measurements were performed at room temperature using a confocal microscope and a single-crystal diamond (Element Six DNV-B14) with dimensions of $3.0\times3.0\times0.5$ mm, $\{100\}$ faces, approximately 4.5 ppm NV centres and natural-abundance $^{13}\mathrm{C}$ (1.1\%). Continuous-wave 532-nm excitation was gated by an acousto-optic modulator and focused through a $100\times$ objective with numerical aperture 0.90. A three-axis piezoelectric stage positioned the sample. Collected fluorescence passed through a 30-µm pinhole and a filter centred at 650 nm with a 10-nm bandwidth before detection by a single-photon counter. The detected rate was typically $2.8$--$3.1\times10^6$ counts s$^{-1}$, with an imposed operating limit of $3.4\times10^6$ counts s$^{-1}$.

The permanent-magnet field was determined by fitting the resonance spectrum of all four NV orientations, giving a field magnitude of 4.18 mT directed $3.8^\circ$ from one NV axis. Microwave excitation addressed the $m_s=0\rightarrow m_s=-1$ transition of this orientation near 2.7503 GHz. The carrier was tuned to the central line of the $^{14}\mathrm{N}$ hyperfine triplet, whose adjacent lines were separated by 2.16 MHz.

An FPGA controller synchronized optical and microwave switching and recorded photon arrivals. Programmed in-phase and quadrature waveforms controlled microwave phase and frequency offset through a mixer. A fast switch blocked microwave transmission between pulses, and an amplifier drove a gold-wire antenna beside the diamond. Waveform offsets were adjusted to suppress carrier leakage, and gain and phase imbalance were corrected using a $2\times2$ matrix, with parameters adjusted individually against spectrum-analyser readings. Qualification measurements compared Rabi oscillations at phases of $0^\circ$, $60^\circ$ and $120^\circ$, and checked residual drive using a zero-amplitude waveform with the microwave switch open. Each new controller program was simulated before execution on the instrument. Microwave pulses in the final calibration and target sequences lasted 64 ns, with 152-ns intervals between pulses.

Fluorescence was recorded during a 50-µs laser pulse following microwave manipulation. Counts from three successive readout windows were combined with fixed weights of 1, 0.45 and 0.10, respectively, to obtain the weighted photon count per shot. Sequence means were expressed in these units. The same laser pulse initialized the spin ensemble for the next shot. A 1-µs dark interval completed an approximately 52-µs shot cycle, and all sequence settings used the same total shot duration.

\subsection{Acquisition protocol and amendments}
A preliminary measurement of the predicted identity used 20-ns inter-pulse gaps and 16 five-minute sub-runs. The first main acquisition assigned equal exposure to each sequence and comprised 36 alternating ten-minute sub-runs. Two controller programs contained complementary sets of global-phase replicas: program A applied common phase shifts of $0^\circ$, $120^\circ$ and $240^\circ$, and program B applied $60^\circ$, $180^\circ$ and $300^\circ$, to all pulses in each nonempty sequence. Following discussion between the manager and PI, this schedule replaced the initially proposed randomized 66-s cycles.

Failures of the global-phase and reference-pulse controls, associated with signal changes sampled at different times by the two programs, prompted amendments to the acquisition. The first run retained its inconclusive classification. A first pilot shortened the alternation to 30-s blocks and used three sequence orders. A second pilot implemented phase alternation within a single controller program, adding $60^\circ$ to every pulse phase on alternate sweeps through the sequences. The pilots were classified as acquisition development, and the original decision criteria were retained.

The amended protocol adopted the within-program phase toggle and specified 720 blocks of 30 s, giving six hours of integration. Three prespecified sequence orders were arranged in 120 randomized six-block groups. Before the final acquisition, the protocol required a 20-minute diagnostic comparing the null pair: the empty sequence and a shot with microwaves disabled. These two settings exchanged positions during the diagnostic to assess their sensitivity to acquisition order. The protocol also required at least 20 min of warm-up and 240 s of operation before recording, while maintaining the laser duty cycle. A signal-level step was to be flagged if it exceeded both 1\% and six standard errors.

A 32-bit integer overflow in the reference-diagnostic calculation produced an undefined standard error. The experimentalist corrected the calculation using floating-point arithmetic and reanalysed the same diagnostic data against the unchanged cutoff. The final acquisition accumulated 21,600 valid one-second records across eight segments, completing the prescribed six hours of integration.

\subsection{Statistical analysis and outcome classification}
For each withheld target, the prediction residual was defined as $r_i=T_i-(H\mathbf C)_i$, where $\mathbf C$ contains the ten calibration means and $H$ is the fixed transfer matrix. The primary accuracy test used ten margins,
\[
S-0.8P,\qquad 1.2P-S,\qquad
0.3P+r_i,\qquad 0.3P-r_i
\quad (i=1,\ldots,4).
\]
Positive margins corresponded to agreement within 20\% of $P$ for the measured contrast and residual magnitudes below 30\% of $P$ for each target.

Standard errors were propagated using the complete coefficients of the underlying weighted-count means. This propagation included uncertainty in the calibration-derived predictions and the covariance between $P$ and each residual arising from their shared calibration data. Poisson-based variance contributions were multiplied by an excess-variance factor $F^*$, with a floor of 1.1. Values of $F^*>1.25$ failed the variance-validity requirement.

Control measurements assessed global-phase invariance and consistency of the null signal. A joint $\chi^2$ statistic compared the six global-phase replica means for each of the 13 nonempty sequences, giving 65 degrees of freedom. A separate phase-invariance check used six replicas of a single 128~ns reference pulse. The null control tested equality between the mean signal of the empty sequence and that of a shot with microwave emission disabled. Each control used a rejection threshold corresponding to a tail probability of $10^{-4}$; the null comparison was two-sided.

Acquisitions were classified using the fixed decision rule. Support required all control and variance-validity criteria to pass, positive individual one-sided 99\% lower confidence bounds for all ten margins, and a positive measured contrast $S$ at one-sided $p\leq0.001$. Contradiction required the same validity criteria, a predicted contrast $P$ more than 3.09 standard errors above zero, and at least one margin with a negative one-sided upper confidence bound at tail probability $10^{-4}$. All remaining cases were classified as inconclusive, including any acquisition that failed a control or the variance-validity requirement.

The primary classification therefore assessed agreement within the predefined tolerances. Later diagnostic analyses examined joint consistency of the residuals with zero and the dependence of $S$ and $P$ on acquisition order. These diagnostics were separate from the fixed outcome-classification rule.

\subsection*{Acknowledgements}
This work was supported by JST FOREST Program (Grant No.~JPMJFR2547). E.K. was supported by JST SPRING (Grant No.~JPMJSP2180).

\subsection*{Author Contributions}
E.M.M. conceived the study, developed the AI-agent framework and the experimental platform, integrated the framework with the experimental platform, analysed the data, and drafted the manuscript. E.K. constructed and developed the experimental platform and contributed to the structure and writing of the manuscript. M.K. constructed and configured the experimental apparatus and reviewed the manuscript. K.A. secured funding, supervised the project, and revised the manuscript. All authors reviewed and approved the final manuscript.

\subsection*{Competing Interests}
The authors declare no competing interests.

\subsection*{AI Use Disclosure}
The authors used generative AI agents for the research methodology, including scientific reasoning, computational analysis, experimental planning and execution, and manuscript drafting. The human authors take full responsibility for the scientific claims, data integrity, and final manuscript.

\subsection*{Data / Code Availability}
The experimental data and analysis code supporting the findings of this study are available from the corresponding author upon reasonable request.

\subsection*{Open-Source Release}
To make the full pipeline openly available, we release Auctopus, an open-source implementation of the multi-agent laboratory described in this work, at \url{https://github.com/E-zClap/auctopus}. It runs every stage of the workflow, from proposal generation to laboratory measurement and review, through the OpenAI and Anthropic APIs. The prompts used in this study are not included; the release provides templates to adapt to other research questions and instruments. Auctopus is distributed under the MIT licence, and we welcome its use, adaptation and contributions to its further development.

\bibliographystyle{unsrtnat}
\bibliography{refs.bib}
\end{document}